# Clustering Mixed Numeric and Categorical Data: A Cluster Ensemble Approach[*]

Zengyou He, Xiaofe i Xu, Shengchun Deng

*Department of Computer Science and Engineering, Harbin Institute of Technology*

*Harbin 150001, P. R. China*

Email: zengyouhe@yahoo.com, xiaofei@hit.edu.cn, dsc@hit.edu.cn

**Abstract** Clustering is a widely used technique in data mining applications for discovering patterns in underlying data. Most traditional clustering algorithms are limited to handling datasets that contain either numeric or categorical attributes. However, datasets with mixed types of attributes are common in real life data mining applications. In this paper, we propose a novel divide-and-conquer technique to solve this problem. First, the original mixed dataset is divided into two sub-datasets: the *pure* categorical dataset and the *pure* numeric dataset. Next, existing well established clustering algorithms designed for different types of datasets are employed to produce corresponding clusters. Last, the clustering results on the categorical and numeric dataset are combined as a categorical dataset, on which the categorical data clustering algorithm is used to get the final clusters. Our contribution in this paper is to provide an algorithm framework for the mixed attributes clustering problem, in which existing clustering algorithms can be easily integrated, the capabilities of different kinds of clustering algorithms and characteristics of different types of datasets could be fully exploited. Comparisons with other clustering algorithms on real life datasets illustrate the superiority of our approach.

**Keywords** Clustering, Mixed Type Attributes, Data Mining, Categorical Data

## 1. Introduction

Clustering typically groups data into sets in such a way that the intra-cluster similarity is maximized while the inter-cluster similarity is minimized. The clustering technique has been extensively studied in many fields such as pattern recognition, customer segmentation, similarity search and trend analysis. The capability to deal with datasets contains both numeric and categorical attributes is undoubtedly important due to the fact that datasets with mixed types of attributes are common in real life data mining applications. Although many clustering algorithms have been proposed so far, most of those algorithms are designed to find clusters on an assumption that all the attributes are either numeric or categorical.

Most previous clustering algorithms focus on numerical data whose inherent geometric properties can be exploited naturally to define distance functions between data points. However, much of the data existed in the databases is categorical, where attribute values can't be naturally ordered as numerical values. An example of categorical attribute is *shape* whose values include *circle*, *rectangle*, *ellipse*, etc. Due to the differences in the characteristics of these two kinds of data, attempts to develop criteria functions for mixed data have been not very successful.

[*] This work was supported by the High Technology Research and Development Program of China (No. 2002AA413310) and the IBM SUR Research Fund.

In this paper, we propose a novel divide-and-conquer technique to solve this problem. First, the original mixed dataset is divided into two sub-datasets: the *pure* categorical dataset and the *pure* numeric dataset. Next, existing well established clustering algorithms designed for different types of datasets are employed to produce corresponding clusters. Last, the clustering results on the categorical and numeric dataset are combined as a categorical dataset, on which the categorical data clustering algorithm is employed to get the final output.

Our main contribution in this paper is to provide an algorithm framework for the mixed attributes clustering problem, in which existing clustering algorithms can be easily integrated, the capabilities of different kinds of clustering algorithms and characteristics of different types of datasets could be fully exploited.

The remainder of this paper is organized as follows. Section 2 has a detailed discussion on related work. Section 3 argues that the cluster ensemble problem can be revisited as the categorical data clustering problem. In Section 4, we describe the cluster ensemble based algorithm framework for clustering mixed categorical and numeric data. Experimental results are given in Section 5 and Section 6 concludes the paper.

## 2. Related Work

### 2.1 Clustering Mixed Data

From the viewpoint of target dataset for analysis, existing clustering algorithms can be classified into three categories: numeric, categorical and mixed.

Most previous clustering algorithms focus on numerical data whose inherent geometric properties can be exploited naturally to define distance functions between data points, such as DBSCAN [3], BIRTH [4], $C^2P$ [2], CURE [5], CHAMELEON [6], WaveCluster [7]. As pointed out in [8], numeric clustering algorithms are not appropriate for categorical attributes; therefore, it is easy to conclude that they are also not suitable for the task of clustering mixed type attributes.

A few algorithms have been proposed in recent years for clustering categorical data [1, 8~14]. However, all of these algorithms are designed for categorical attributes and their extending capability to mixed type attributes are unknown.

For the problem of clustering mixed type attributes, some research efforts have been done [15~17]. In [15], the author presents two algorithms, *k*-modes and *k*-prototypes, which extend the *k*-means paradigm to categorical domains and domains with mixed attributes. A new distance measure for categorical attributes based on the total mismatches of the categorical attributes of two data records is proposed in the *k*-modes algorithm. For the mixed type of attributes, the *k*-prototypes algorithm used a weighted sum of Euclidean distance for numeric attributes and the proposed distance measure for categorical attributes. However, the weights have to be determined a priori. Improper weights may result in biased treatment of different attributes types.

In [16], the SBAC algorithm is proposed, which adopts a similarity measure that gives greater weight to uncommon feature value matches in similarity computations and employs an agglomerative algorithm to construct a dendrogram. However, for the complexity of the SBAC algorithm is quadratic in the number of records in the dataset, it is almost impossible to handle very large datasets.

In [17], the authors introduce their clustering algorithm that is available commercially in the

Clementine 6.0 data mining tool. Their distance measure is derived from a probabilistic model that the distance between two clusters is equivalent to the decrease in log-likelihood function as a result of merging. Their clustering algorithm is based on the framework of BIRTH [4] using proposed distance measure. The BIRTH algorithm has the drawback that it may not work well when clusters are not "spherical and it can be affected by the input order of records. Thus, the same problems exist for the clustering algorithm of [17].

## 2.2  Cluster Ensembles

Cluster ensembles is the method to combine several runs of different clustering algorithms to get a common partition of the original dataset, aiming for consolidation of results from a portfolio of individual clustering results. Although the research on cluster ensembles has not been widely recognized as that combing multiple classifier or regression models, more recently, several research efforts have been done independently [18~20].

In [18], the authors formally defined the cluster ensemble problem as an optimization problem and propose combiners for solving it based on a hyper-graph model.

A multi-clustering fusion method is presented in [19]. In their method, the results of several independent runs of the same clustering algorithm are appropriately combined to obtain a partition of the data that is not affected by initialization and overcomes the instabilities of clustering methods. After that, the fusion procedure starts with the clusters produced by the combining part and finds the optimal number of clusters according to some predefined criteria.

The authors in [20] proposed a sequential combination method to improve the clustering performance. First, their algorithm uses the global criteria based clustering to produce an initial result, then use the local criteria based information to improve the initial result with a probabilistic relaxation algorithm or linear additive model.

## 3.  Cluster Ensemble: The Viewpoint of Categorical Data Clustering

Our goals in this section are twofold: 1) to argue that the cluster ensemble problem can be revisited as the categorical data clustering problem and 2) illustrate that existing cluster combination methods are not appropriate in this literature.

Clustering aims at discovering groups and identifying interesting distribution and patterns in data sets. In general, the output produced by a special clustering algorithm will be the assignment of data objects in dataset to different groups. In other words, it will be sufficient to identify each data object with a unique cluster label. From the viewpoint of clustering, data objects with different cluster labels are considered to be in different clusters, if two objects are in the same cluster then they are considered to be fully similar, if not they are fully dissimilar. Thus, it is obvious that cluster labels are impossible to be given a natural ordering in a way similar to real numbers, that is to say, the output of clustering algorithm can be viewed as *categorical* (or *nominal*).

Cluster ensemble is the method to combine several runs of different clustering algorithms to get a common partition of the original dataset, aiming for consolidation of results from a portfolio of individual clustering results. Since the output of individual clustering algorithm is categorical and so the cluster ensemble problem can viewed as the categorical data clustering problem, in which runs of different clustering algorithm are combined into a new categorical dataset.

Transforming the cluster ensemble problem into the categorical data clustering problem has the following advantages. First, some efficient algorithms for clustering categorical data have been proposed recently [1, 8~14]. These algorithms can be fully exploited and also cluster ensemble problem could benefit from the advances in the research of categorical data clustering. Further, the problem of categorical data clustering is relatively simple and provides a unified framework for problem formalization.

For clustering datasets with mixed types of attributes, we propose a novel divide-and-conquer technique. First, the original mixed dataset is divided into two sub-datasets: the *pure* categorical dataset and the *pure* numeric dataset. Next, existing well established clustering algorithms designed for different types of datasets are employed to produce corresponding clusters. Last, the clustering results on the categorical and numeric dataset are combined as a categorical dataset, on which the categorical data clustering algorithm is exploited to get the final clusters.

Now we focus on the last step of our technique. In fact, with the clustering results on the categorical and numeric dataset, we can get the final clustering results by exploiting existing cluster ensemble methods [18~20]. However, existing cluster combination methods have some limitations on their own ways and are not appropriate for our problem.

Assuming that the dataset with mixed attributes has $m$ attributes, with $m_c$ categorical attributes and $m_n$ numeric attributes, $m_c + m_n = m$. And each attribute has equal importance, that is, all the attributes have the same weights and no biased treatment to any attributes. It is often the case that $m_c \neq m_n$, therefore, we must assign $m_c$ and $m_n$ as the *weights* for the clustering output on categorical dataset and numeric dataset when performing combination or categorical data clustering.

Unfortunately, the approach in [19] is designed for combining runs of clustering algorithms with the same number of clusters and the ability to handle such *weighted cluster ensemble* problem is unknown. The sequential combination method proposed in [20] has the same problem as the approach in [19]. In addition, their algorithm has the limitation to combine only the outputs of *two specific* clustering algorithms. The authors in [18] proposed combiners based on a hyper-graph model to solve the cluster ensemble problem. Their algorithms are easy to be extended to fit the *weighted cluster ensemble* problem, however, all of these algorithms involve high computation costs, which is not feasible in the environment of access to large data sets stored in secondary memory. Furthermore, their effectiveness in cluster ensemble is not fully verified.

Thus, in this paper, we use the categorical data clustering approach to handle the cluster ensemble problem.

## 4. Cluster Ensemble Based Algorithm

In this section, we describe the cluster ensemble based algorithm framework for clustering mixed categorical and numeric data. We begin by presenting overview of the algorithm framework, and reserve the details and complexity results for subsequent sub-sections.

## 4.1 Overview

The steps involved in the cluster ensemble based algorithm framework are described in Figure 1. First, the original mixed dataset is divided into two sub-datasets: the *pure* categorical dataset and the *pure* numeric dataset. Next, existing well established clustering algorithms designed for different types of datasets are employed to produce corresponding clusters. Finally, the clustering results on the categorical and numeric dataset are combined as a categorical dataset, on which the categorical data clustering algorithm is exploited to get the final clusters.

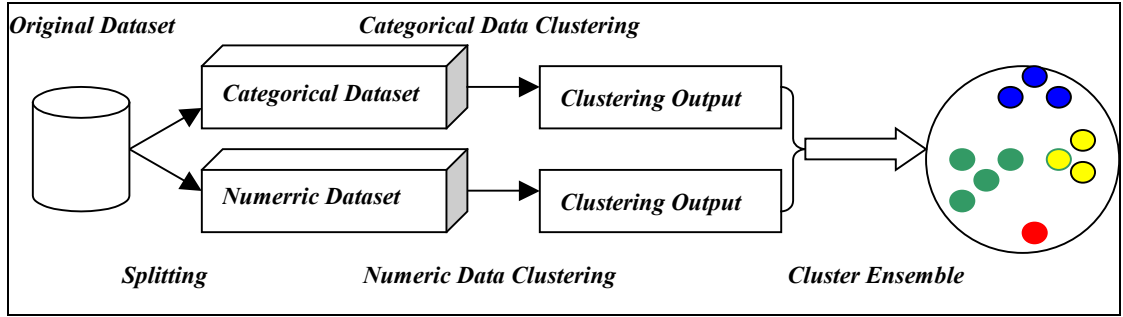

**Fig.1.** Overview of CEBMDC algorithm framework

For this algorithm framework gets clustering output from both splitting categorical dataset and numeric dataset, therefore, it is named as CEBMDC (**C**luster **E**nsemble **B**ased **M**ixed **D**ata **C**lustering).

## 4.2 The Weighted *Squeezer* Algorithm

In Section 3, we argue that using the categorical data clustering approach to handle the cluster ensemble problem is more appropriate in our problem. Therefore, we need a categorical data clustering algorithm first. In this paper, the clustering algorithm used for categorical data clustering and cluster ensemble is the *Squeezer* algorithm [1], which can produce good clustering results and at the same time deserves good salability. Next, we will give an introduction about *Squeezer* algorithm.

Let $A_1, ..., A_m$ be a set of categorical attributes with domains $D_1, ..., D_m$ respectively. Let the dataset $D$ be a set of tuples where each tuple $t$: $t \in D_1 \times ... \times D_m$. Let *TID* be the set of unique ID of every tuple. For each $tid \in TID$, the attribute value for $A_i$ of corresponding tuple is represented as $tid.A_i$.

**Definition 1:** (*Cluster*) *Cluster* = {$tid$ | $tid \in TID$} is subset of *TID*.

**Definition 2:** Given a *Cluster C*, the set of different attribute values on $A_i$ with respect to $C$ is defined as: $VAL_i(C) = \{tid.A_i \mid tid \in C\}$ where $1 \leq i \leq m$.

**Definition 3:** Given a *Cluster C*, let $a_i \in D_i$, the support of $a_i$ in $C$ with respect to $A_i$ is defined as: $Sup(a_i) = |\{tid \mid tid.A_i = a_i\}|$.

**Definition 4:** (*Summary*) Given a *Cluster C*, the *Summary* for $C$ is defined as:

$$Summary = \{VS_i \mid 1 \leq i \leq m\} \text{ where } VS_i = \{(a_i, Sup(a_i)) \mid a_i \in VAL_i(C)\}.$$

Intuitively, the *Summary* of a *Cluster* contains summary information about this *Cluster*. In general, each *Summary* will consists of $m$ elements, where $m$ is number of attributes. The element in *Summary* is the set of pairs of attribute values and their corresponding supports. We will show later that information contained in *Summary* is sufficient enough to compute the similarity between a tuple and *Cluster*.

**Definition 5:** (*Cluster Structure, CS*) Given a *Cluster C*, the *Cluster Structure (CS)* for *C* is defined as: *CS = {Cluster, Summary}*.

**Definition 6:** Given a *Cluster C* and a tuple *t* with *tid* ∈ *TID*, the *similarity* between *C* and *tid* is defined as:

$$Sim(C, tid) = \sum_{i=1}^{m} w_i \left( \frac{Sup(a_i)}{\sum_{a_j \in VAL_i(C)} Sup(a_j)} \right) \text{ where } tid.A_i = a_i \text{ and } w_i \text{ is the } weight \text{ of attribute } A_i.$$

From the definition 6, it is clear that the similarity used here is capable to handle clustering problem with different weights on different attributes. That is, it is appropriate for *weighted cluster ensemble* problem. In the *Squeezer* algorithm, this measure is used to determine weather the tuple should be put into the cluster or not.

---

**Algorithm** *Squeezer* (*D, s*)
**Begin**
1. while (*D* has unread tuple){
2.    tuple = getCurrentTuple (*D*)
3.    if (tuple.tid == 1){
4.      addNewClusterStructure (tuple.tid) }
5.    else{
6.      for each existed cluster *C*
7.        simComputation(C,tuple)
8.      get the max value of similarity: *sim_max*
9.      get the corresponding Cluster Index: *index*
10.     if *sim_max* >= *s*
11.       addTupleToCluster(tuple, index)
12.     else
13.       addNewClusterStructure (tuple.tid) }
14. }
15. outputClusteringResult ()
**End**

**Fig. 2.** *Squeezer* Algorithm

---

The *Squeezer* algorithm has *n* tuples as input and produce clusters as final results. Initially, the first tuple in the database is read in and a *Cluster Structure (CS)* is constructed with *C* ={1}. Then, the consequent tuples are read iteratively.

For every tuple, by the similarity function, we compute its similarities with all existing clusters, which are represented and embodied in the corresponding *CSs*. The largest value of similarity is selected out. If it is larger than the given threshold, donated as *s*, the tuple will be put into the cluster that has the largest value of similarity. The *CS* is also updated with the new tuple. If the above condition does not hold, a new cluster must be created with this tuple.

The algorithm continues until it has traversed all the tuples in the dataset. It is obvious that the *Squeezer* algorithm only makes one scan over the dataset, thus, highly efficient for disk resident datasets where the I/O cost becomes the bottleneck of efficiency.

The *Squeezer* algorithm is presented in Figure 2. It accepts as input the dataset *D* and the value of the desired similarity threshold. The algorithm fetches tuples from *D* iteratively.

Initially, the first tuple is read in, and the sub-function *addNewClusterStructure()* is used to establish a new *Clustering Structure*, which includes *Summary* and *Cluster* (Step 3-4).

For the consequent tuples, the similarity between an existed *Cluster C* and each tuple is computed using sub-function *simComputation()* (Step 6-7). We get the maximal value of similarity (donated by *sim_max*) and the corresponding index of *Cluster* (donated by *index*) from the above computing results (Step 8-9). Then, if the *sim_max* is larger than the input threshold *s*, sub-function *addTupleToCluster()* will be called to assign the tuple to selected *Cluster* (Step 10 11). If it is not the case, the sub-function *addNewClusterStructure()* will be called to construct a new *CS* (Step 12-13). Finally, the clustering results will be labeled on the disk (Step 15).

Choosing the *Squeezer* algorithm as the background clustering algorithm for categorical data clustering and cluster ensemble in this literature is based on the consideration that this algorithm has the following novel features:
- ✓ It achieves both high quality of clustering results and scalability.
- ✓ Its ability for handling high dimensional datasets effectively.

## 4.3  The Algorithm and Computation Complexity

In this section, we describe the algorithm based on CEBMDC framework (named *algCEBMDC*), which is described in Figure 3.

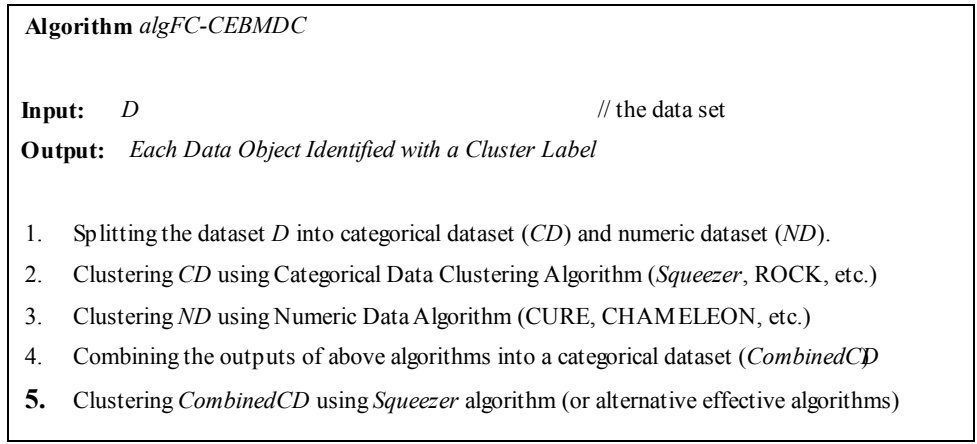

**Algorithm** *algFC-CEBMDC*

**Input:** *D*           // the data set
**Output:** *Each Data Object Identified with a Cluster Label*

1. Splitting the dataset *D* into categorical dataset (*CD*) and numeric dataset (*ND*).
2. Clustering *CD* using Categorical Data Clustering Algorithm (*Squeezer*, ROCK, etc.)
3. Clustering *ND* using Numeric Data Algorithm (CURE, CHAMELEON, etc.)
4. Combining the outputs of above algorithms into a categorical dataset (*CombinedCD*)
5. Clustering *CombinedCD* using *Squeezer* algorithm (or alternative effective algorithms)

**Fig. 3.** The *algCEBMDC* Algorithm

The computational complexity of the *algCEBMDC* algorithm is derived in three parts: 1) the complexity for clustering the categorical dataset, 2) the complexity for clustering the numeric dataset and 3) the complexity for clustering the combined categorical dataset. Therefore, the overall complexity for the *algFC-CEBMDC* algorithm is O ($C_1+C_2+C_3$), where $C_1$, $C_2$ and $C_3$ are the computation complexities of part 1), part 2) and part 3).

One observation from the above analysis is that our algorithm's computation complexity is determined by the component clustering algorithms. So far, many efficient clustering algorithms for large databases are available, it implicate that our algorithms will effective for large-scale data mining applications, too.

## 5. Experimental Results

A comprehensive performance study has been conducted to evaluate our algorithm. In this section, we describe those experiments and their results. We ran our algorithm on real-life datasets obtained from the UCI Machine Learning Repository [21] to test its clustering performance against other algorithms. At the same time, its properties are also empirically studied.

### 5.1 Experiments Setup

In our current implementation, the clustering algorithm used for numeric dataset is provided in the CLUTO software package [22]. CLUTO is a software package for clustering low- and high-dimensional datasets and for analyzing the characteristics of the various clusters. It provides three different classes of clustering algorithms that operate either directly in the object's feature space or in the object's similarity space. These algorithms are based on the partitional, agglomerative, and graph-partitioning paradigms. In our experiments, we use the graph-partitioning based algorithm for clustering numeric dataset, which is a variation of the CHAMELEON [6] algorithm.

In the *algCEBMDC* algorithm, the *Squeezer* algorithm is used for both clustering categorical dataset and performing cluster ensemble.

### 5.2 Real Life Datasets

Our experiments with real data sets focus on the comparing the quality of the clustering results produced by our algorithm with other algorithms, such as *k*-prototypes. We choose datasets based on the consideration that they contain approximate equal numbers of categorical attributes and numeric attributes.

The first dataset was the credit approval dataset. The dataset has 690 instances, each being described by 6 numeric and 9 categorical attributes. The instances were classified into two classes, approved labeled as "+" and rejected labeled as "−". We compare our algorithm with *k*-prototypes algorithm; however, the *k*-prototypes algorithm cannot handle missing values in numeric attributes, 24 instances with missing values in numeric attributes were removed. Therefore, only 666 instances were used.

The second dataset was *cleve* dataset, which is Dr. Detrano's heart disease dataset that generated at the Cleveland Clinic[1] modified to be a real mixed dataset. The dataset has 303 instances, each being described by 6 numeric and 8 categorical attributes. The instances were also classified into two classes, each class is either healthy (buff) or with heart-disease (sick). The

---

[1] The Cleveland Clinic heart disease database is maintained at the UCI repository for machine learning databases. Acknowledgment goes to Robert Detrano, MD, PhD as the principle investigator for the data collection study.

*cleve* dataset has 5 missing values in numeric attributes, all of them are replaced with the value "0".

The clustering accuracy for measuring the clustering results is computed as the follows, which has been done in [15]. Suppose the final number of clusters is *k*, clustering accuracy *r* is defined as: $r = \frac{\sum_{i=1}^{k} a_i}{n}$, where *n* is number of instances in the dataset, $a_i$ is number of instances occurring in both cluster *i* and its corresponding class, which has the maximal value. In other words, $a_i$ is the number of instances with class label that dominate cluster *i*. Consequently, the clustering error is defined as $e = 1 - r$.

## 5.3 Clustering Results

We used the *algCEBMDC* and *k*-prototypes algorithms to cluster the credit approval dataset and the *cleve* dataset into different numbers of clusters, varying from 2 to 9. For each fixed number of clusters, the clustering errors of different algorithms were compared.

For both of datasets, the *k*-prototypes algorithm, just as has been done in [15], all numeric attributes are rescaled to the range of [0.1]. On the credit approval dataset, the *algCEBMDC* algorithm let CLUTO software generate 4 clusters on the splitting numeric dataset and let the *Squeezer* algorithm produce 4 clusters on the splitting categorical dataset with the similarity threshold set to 4.

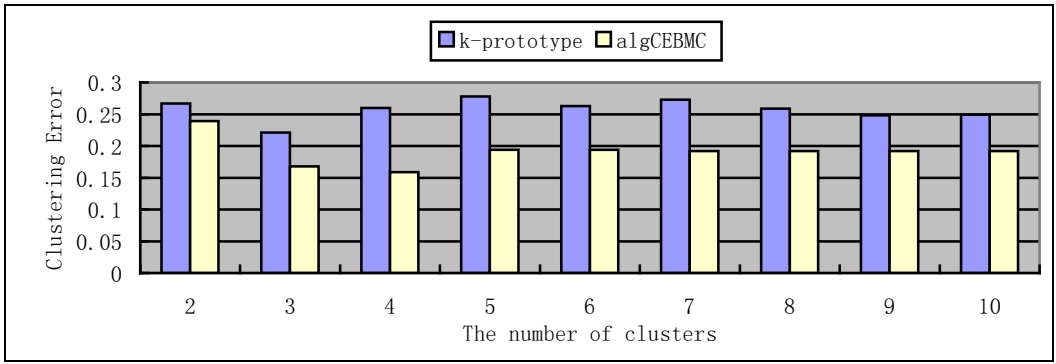

**Fig.4.** Clustering error vs. Different number of clusters (credit approval dataset)

Figure 4 shows the results on the credit approval dataset of different clustering algorithms. From Figure 4, we can summarize the relative performance of our algorithms as follows (See Table 1):

**Table 1:** Relative performance of different clustering algorithms (credit approval dataset)

| Ranking | 1 | 2 | Average Clustering Error |
|---|---|---|---|
| ***k*-prototypes** | 0 | 9 | 0.258 |
| ***algCEBMC*** | 9 | 0 | 0.191 |

That is, comparing with the *k*-prototypes algorithm, our algorithm performed the best in all cases. It never performed the worst. Furthermore, the average clustering errors of our algorithm are significantly smaller than that of the *k*-prototypes algorithm.

On the *cleve* dataset, the *algCEBMDC* algorithm let CLUTO software generate 4 clusters on the splitting numeric dataset and let the *Squeezer* algorithm produce 3 clusters on the splitting categorical dataset with the similarity threshold set to 3. The experimental results on the *cleve* dataset are described in Figure 5 and the summarization on the relative performance of the 2 algorithms is given in Table 2.

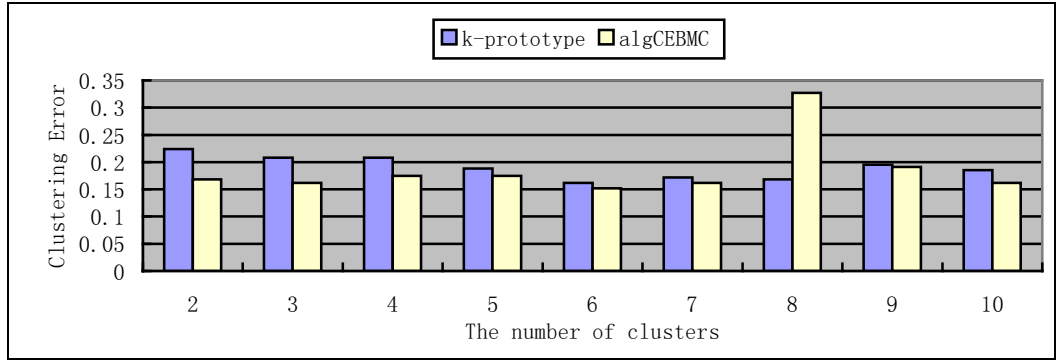

**Fig.5.** Clustering error vs. Different number of clusters (*cleve* dataset)

From Figure 5 and Table 4, although the average clustering accuracy of our algorithm is only a little better than that of the *k*-prototypes algorithm, while the cases of our algorithms that beat the *k*-prototypes algorithm is dominant in this experiment. Therefore, the test on the *cleve* dataset also verified the superiority of our method.

**Table 2:** Relative performance of different clustering algorithms (*cleve* dataset)

| Ranking | 1 | 2 | Average Clustering Error |
|---|---|---|---|
| ***k*-prototypes** | 1 | 8 | 0.190 |
| ***algCEBMC*** | 8 | 1 | 0.187 |

The above experimental results demonstrate the effectiveness of *algCEBMC* algorithm for clustering dataset with mixed attributes. In addition, it outperforms the *k*-prototypes algorithm with respect to clustering accuracy.

## 5.4 Properties

In this section, we empirically study some interesting properties of the *algCEBMC* algorithm. The first test aims to find how the clustering accuracy and the final number of clusters changed in the cluster ensemble stage with different parameters. The datasets used are still the credit card approval dataset and the *cleve* dataset, the clusters generated on the splitting numeric dataset and the splitting categorical dataset are just the same as done in Section 5.3. Since the only parameter needed by the *Squeezer* algorithm is the similarity threshold, we normalized it to the range of [0.1].

Figure 6 and Figure 7 show how the clustering accuracy and the final number of clusters changed with different similarity thresholds in the cluster ensemble stage. From these two figures, it can be observed that, with the fixed clustering output on the splitting categorical dataset and numeric dataset, the final clustering results of our approach are very stable, because the parameter in a wide range gives the same results. That is, in the cluster ensemble stage, the final output is parameter insensitive.

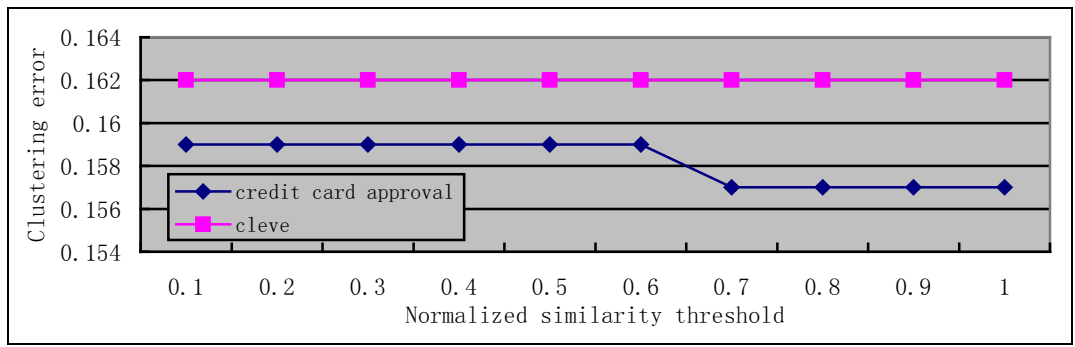

**Fig.6.** Clustering error vs. Different value of similarity threshold in the cluster ensemble stage

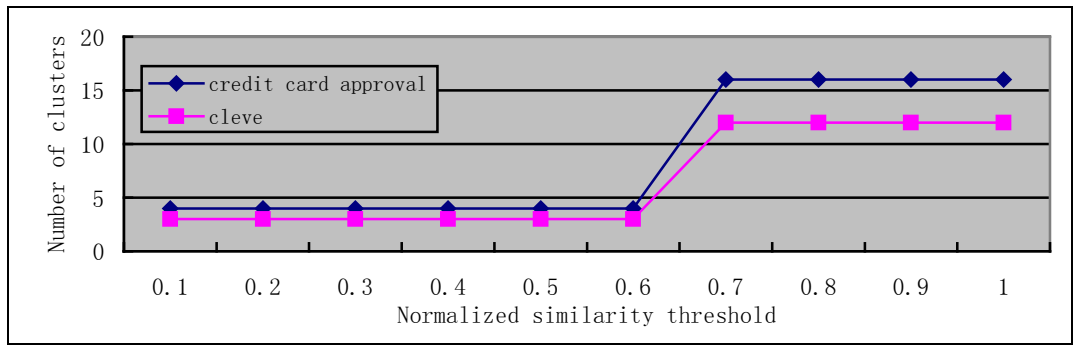

**Fig.7.** Number of clusters vs. Different value of similarity threshold in the cluster ensemble stage

The second test goes to how the underlying outputs on the splitting categorical dataset and numeric dataset affect the final clustering results. We use $k_c$ donate the number of clusters on the categorical dataset and $k_n$ for that of numeric dataset. And $k$ is used to donate the final number of clusters. In this test, we let $k = k_c = k_n$ and their values varied from 2 to 10, then the clustering accuracies of the final output, the output on categorical dataset and the output on numeric dataset are compared. Figure 8 and Figure 9 describe the results on the credit approval dataset and the *cleve* dataset.

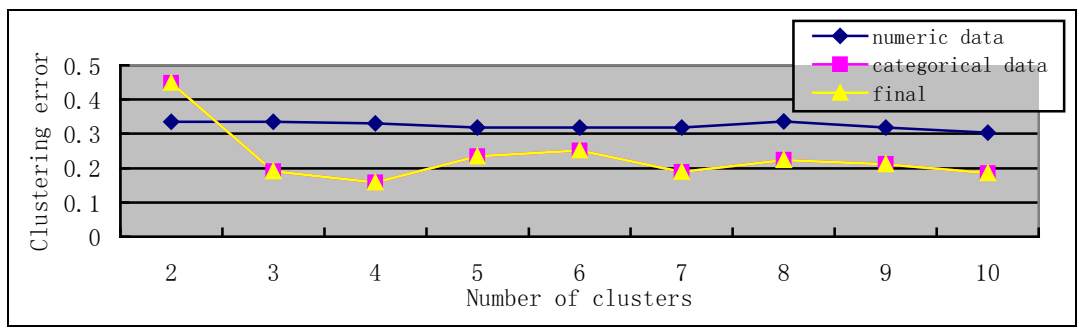

**Fig.8.** Clustering error vs. Different number of clusters (credit approval dataset)

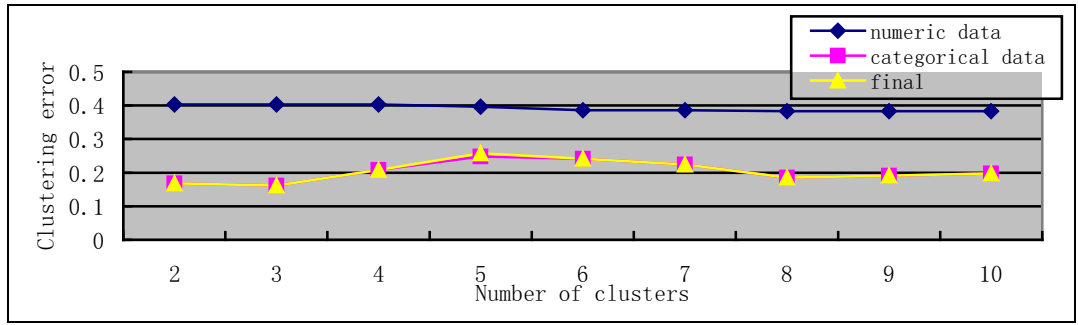

**Fig.9.** Clustering error vs. Different number of clusters (*cleve* dataset)

One important observation from the Figure 8 and Figure 9 is that the clustering accuracy of *algCEBMC* algorithm (final) is determined by the underlying outputs on the splitting categorical dataset and numeric dataset. For in the cluster ensemble stage, the separate clustering results are the basis for combination, therefore, the quality of clustering outputs on the two kinds of datasets is critical to our algorithm's clustering accuracy. That is, effective categorical data clustering algorithms and numeric data clustering algorithms should be selected as the component algorithms for generating clusters.

Another observation is that the clustering accuracy of *algCEBMC* algorithm (final) is almost identical to that of the categorical dataset. This phenomenon happened because our test datasets contain more categorical attributes than numeric attributes. The categorical datasets dominate the data distribution in both datasets.

## 6. Conclusions

Existing solutions for clustering mixed numeric and categorical data fall into the following categories [16]: 1) Encode nominal attribute values as numeric integer values and apply distance measures used in numeric clustering for computing proximity between object pairs. 2) Discretize numeric attributes and apply categorical clustering algorithms. 3) Generalize criterion functions designed for one type of features to handle numeric and nonnumeric feature values.

In this paper, we propose a novel divide-and-conquer technique to solve this problem. First, the original mixed dataset is divided into two sub datasets: the *pure* categorical dataset and the *pure* numeric dataset. Next, existing well established clustering algorithms designed for different types of datasets could be employed to produce corresponding clusters. Finally, the clustering results on the categorical and numeric dataset are combined as a categorical dataset, on which the categorical data clustering algorithm is exploited to get the final clusters.

Our main contribution in this paper is to provide an algorithm framework for the mixed attributes clustering problem, in which existing clustering algorithms can be easily integrated, the capabilities of different kinds of clustering algorithms and characteristics of different types of datasets could be fully exploited.

In the future work, we will investigate integrating other alternative clustering algorithms into the algorithm framework, to get further insight into this methodology. Moreover, applying the proposed divide-and-conquer technique for detecting cluster based local outliers [23] in large database environment with mixed type attributes will be also addressed.

## Acknowledgements

We would like to express our gratitude for Dr. Joshua Huang in the University of Hong Kong for sending us helpful materials and the source codes of *k*-prototypes.